# Learning Equivalence Classes of Bayesian Network Structures

**David Maxwell Chickering**
Computer Science Department
University of California at Los Angeles
*dmax@cs.ucla.edu*

## Abstract

Approaches to learning Bayesian networks from data typically combine a scoring function with a heuristic search procedure. Given a Bayesian network structure, many of the scoring functions derived in the literature return a score for the entire equivalence class to which the structure belongs. When using such a scoring function, it is appropriate for the heuristic search algorithm to search over equivalence classes of Bayesian networks as opposed to individual structures. We present the general formulation of a search space for which the states of the search correspond to equivalence classes of Bayesian networks. Using this space, any one of a number of heuristic search algorithms can easily be applied. We compare greedy search performance in the proposed search space to greedy search performance in a search space for which the states correspond to individual Bayesian network structures.

## 1   INTRODUCTION

Recently, many researchers have developed methods for learning Bayesian networks from data. The available techniques include Bayesian methods [Cooper and Herskovitz, 1991, Buntine, 1991, Spiegelhalter et al., 1993, Heckerman et al., 1995], quasi-Bayesian methods [Lam and Bacchus, 1993, Bouckaert, 1993], and non-Bayesian methods [Pearl and Verma, 1991, Spirtes et al., 1993]. Much of the work in learning Bayesian networks has been devoted to the derivation of a *scoring function*. Given a candidate Bayesian network structure, the scoring function evaluates how well the structure "fits" the observed data and prior knowledge. Once the scoring function has been defined, learning Bayesian networks reduces to a search for one or more structures that have a high score. Chickering (1995a) shows that this search problem is NP-hard when the Bayesian scoring function derived by Heckerman et al. (1995) is used. Consequently, it is ap-propriate to apply heuristic search algorithms in this domain.

Before any search algorithm can be applied to a problem, we must define the three components of a *search space*. First, we need to identify the states of the search, or equivalently, the set of all potential solutions to the search problem. Second, we need a representation for the states of the search. Third, we need a set of operators that transform the representation of one state to another so that the algorithm can traverse the space in a systematic way. Once the search space has been defined, any one of a number of well-known search algorithms can easily be applied to that space.

In perhaps the simplest formulation of a search space for learning Bayesian networks, the states of the search are defined to be individual Bayesian networks, the representation of a state is simply an acyclic directed graph, and the operators are defined to be local changes to those graphs. For example, Chickering et al. (1995) compare various search procedures in the search space of Bayesian networks, using the following operators: for any pair of nodes $x$ and $y$, if $x$ and $y$ are adjacent, the edge connecting them can be either deleted or reversed. If $x$ and $y$ are not adjacent, an edge can be added in either direction. All operators are subject to the constraint that a cycle cannot be formed. We shall use *Bayesian-network space*, or *B-space* for short, to denote the search space defined in this way. Figure 1 shows an example of each operator in B-space.

When two Bayesian networks can represent the same set of probability distributions, we say that those networks are *equivalent*. The relation of network equivalence imposes a set of equivalence classes over Bayesian network structures. Chickering (1995b) shows that many of the scoring functions derived in the literature actually return the measure of fit for the entire equivalence class to which the candidate structure belongs. We call any such scoring function *score equivalent*.

If B-space is used by a search algorithm in conjunction with a score-equivalent scoring function, any particular equivalence class may be represented by a huge number of search states. Furthermore, many of the operators defined for this space move between states corresponding to the same equivalence class. Because the objective of the search algorithm is to identify the state with the highest score, it seems natural that the states of the search space should correspond to equivalence classes of Bayesian networks whenever a score-equivalent scoring function is used.

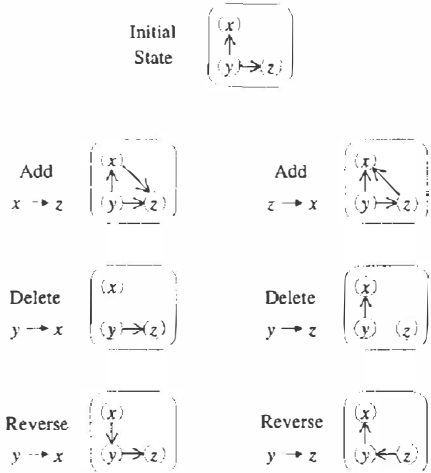

Figure 1: States resulting from the application of a single operator in B-space.

Both Spirtes and Meek (1995) and Madigan et al. (1996) have developed search algorithms that move between equivalence classes as opposed to individual network structures. The (implicit) search spaces that result are well suited for each particular search algorithm, but it is not clear how other search algorithms can efficiently traverse these spaces.

In this paper, we provide the specification for a general search space in which the states are defined to be equivalence classes of Bayesian networks, and to which we can easily apply any one of a number of search algorithms. In Section 2, we introduce our notation and describe previous relevant work. In Section 3, we define the search space. In particular, we describe an efficient representation for equivalence classes, and introduce a set of simple operators that can be applied to this representation. In Section 4, we compare greedy search performance in B-space to greedy search performance in the search space that we define in Section 3. We emphasize that the search space we propose is suitable for any number of search algorithms, and that we have chosen to compare this space to B-space using greedy search because greedy search is very prevalent in the literature on learning Bayesian networks.

## 2   BACKGROUND

In this section, we introduce our notation and describe previous relevant work.

For the remainder of this paper, we will concentrate on Bayesian networks defined over discrete variables. Furthermore, we will assume that the conditional probability distributions are not constrained to belong to any particular functional form.

A Bayesian network $B$ for a set of discrete variables $U = \{x_1, \ldots, x_n\}$ is a pair $(\mathcal{G}, \theta_\mathcal{G})$ where $\mathcal{G} = (U, E_\mathcal{G})$ is a dag, and $\theta_\mathcal{G}$ is the set of conditional probability distributions that correspond to $\mathcal{G}$. We now present a formal definition of equivalence.

**Definition**  *Two dags $\mathcal{G}$ and $\mathcal{G}'$ are* equivalent *if for every Bayesian network $B = (\mathcal{G}, \theta_\mathcal{G})$, there exists a Bayesian network $B' = (\mathcal{G}', \theta_{\mathcal{G}'})$ such that $B$ and $B'$ define the same probability distribution, and vice versa.*

We use $\mathcal{G} \approx \mathcal{G}'$ to denote that $\mathcal{G}$ and $\mathcal{G}'$ are equivalent. As was stated earlier, the relation $\approx$ defines a set of equivalence classes over the network structures. A directed edge $x_i \rightarrow x_j \in E_\mathcal{G}$ is *compelled* in $\mathcal{G}$ if for every dag $\mathcal{G}' \approx \mathcal{G}$, $x_i \rightarrow x_j \in E_{\mathcal{G}'}$. For any edge $e \in E_\mathcal{G}$, if $e$ is not compelled in $\mathcal{G}$, then $e$ is *reversible* in $\mathcal{G}$, that is, there exists some dag $\mathcal{G}'$ equivalent to $\mathcal{G}$ in which $e$ has opposite orientation.

The *skeleton* of any dag is the undirected graph resulting from ignoring the directionality of every edge. A *v-structure* in a dag $\mathcal{G}$ is an ordered triple of nodes $(x, y, z)$ such that (1) $\mathcal{G}$ contains the arcs $x \rightarrow y$ and $z \rightarrow y$, and (2) $x$ and $z$ are not adjacent in $\mathcal{G}$. Verma and Pearl (1990) derive the following characterization of equivalent structures:

**Theorem 1** [Verma and Pearl, 1990] *Two dags are equivalent if and only if they have the same skeletons and the same v-structures.*

Figure 2 shows the 25 states of B-space defined over three-node Bayesian networks. All dags contained within the same dashed rectangle are equivalent – there are 11 equivalence classes for this space.

A consequence of Theorem 1 is that for any edge $e$ participating in a v-structure in some dag $\mathcal{G}$, if that edge is reversed in some other dag $\mathcal{G}'$, then $\mathcal{G}$ and $\mathcal{G}'$ are not equivalent. Thus any edge participating in a v-structure is compelled. Not every compelled edge, however, necessarily participates in a v-structure. For example, the edge from $z$ to $w$ is compelled in the dag shown in Figure 3.

Acyclic partially directed graphs, or *pdags* for short, are graphs that contain both directed and undirected edges, and are commonly used to represent equivalence classes of Bayesian networks.[1] Let $\mathcal{P}$ denote an arbitrary pdag. We define the equivalence class of dags corresponding to $\mathcal{P}$—denoted $Class(\mathcal{P})$—as follows:

---

[1] These graphs are called *patterns* by Spirtes et al. (1993).

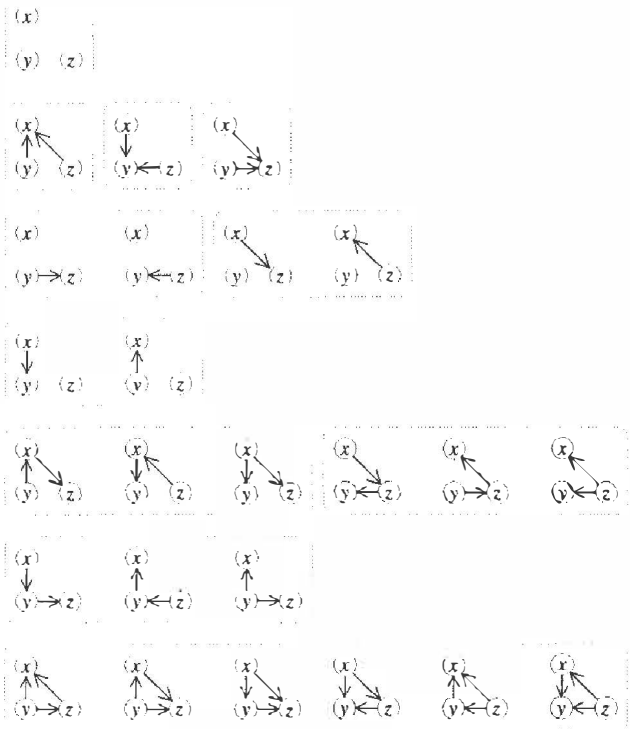

Figure 2: The 25 states of B-space for networks containing three nodes. The equivalent networks in this space are grouped together with a dashed rectangle.

$\mathcal{G} \in Class(\mathcal{P})$ if and only if $\mathcal{G}$ and $\mathcal{P}$ have the same skeleton and the same set of v-structures.[2] From Theorem 1, it follows that a pdag containing a directed edge for every edge participating in a v-structure, and an undirected edge for every other edge, uniquely identifies an equivalence class of dags. There may be many other pdags, however, that correspond to the same equivalence class. For example, any dag interpreted as a pdag can be used with our definition of $Class$ to represent its own equivalence class. It follows that for an equivalence class containing $k$ dags, there are *at least* $k$ different pdags that can represent that class. In the next section, we introduce a subclass of pdags that have a one to one correspondence with equivalence classes.

If a dag $\mathcal{G}$ has the same skeleton and the same set of v-structures as a pdag $\mathcal{P}$, *and* if every directed edge in $\mathcal{P}$ has the same orientation in $\mathcal{G}$, we say that $\mathcal{G}$ is a *consistent extension* of $\mathcal{P}$. Note that any dag that is a consistent extension of $\mathcal{P}$ must also be contained in $Class(\mathcal{P})$, but not every dag in $Class(\mathcal{P})$ is a consistent extension of $\mathcal{P}$. If there is at least one consistent extension of a pdag $\mathcal{P}$, we say that $\mathcal{P}$ *admits a consistent extension*. Figure 4a shows a pdag that admits a consistent extension, and Figure 4b shows one such consistent extension. Figure 4c shows a pdag that does not admit a consistent extension.

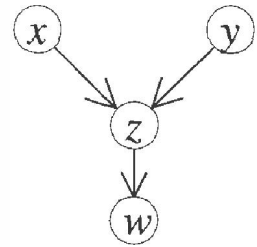

Figure 3: Example of a dag containing a compelled edge that does not participate in a v-structure.

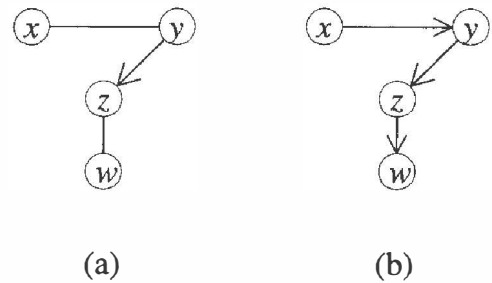

(a)                    (b)

(c)

Figure 4: (a) a pdag that admits a consistent extension, (b) a consistent extension of the pdag in (a) and (c) a pdag that does not admit a consistent extension.

## 3  DEFINING THE SEARCH SPACE

As was discussed in Section 1, a search space has three components: (1) a set of states, (2) a representation for the states, and (3) a set of operators. Given that a score-equivalent scoring function is being used, we have argued that the states of the search should be equivalence classes of Bayesian network structures. In this section, we define an efficient representation for equivalence classes, and a simple set of operators that can be applied to this representation. We call the resulting search space *equivalence-class space*, or *E-space* for short.

Recall the definition of compelled and reversible edges given in Section 2. Every dag in a particular equivalence class has the same set of compelled and reversible edges. Consequently, we can associate compelled and reversible edges with equivalence classes as well as with dags. We define the *completed pdag representation* of an equivalence class to be the pdag consisting of a directed edge for every compelled edge in the equivalence class, and an undirected edge for every reversible edge in the equivalence class.[3] Figure 5a shows a dag $\mathcal{G}$, and Figure 5b shows the completed pdag representation for $Class(\mathcal{G})$.

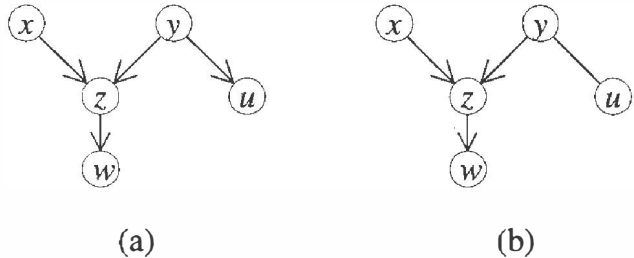

Figure 5: (a) a dag $\mathcal{G}$ and (b) the completed pdag representation for $Class(\mathcal{G})$.

We use completed pdags to represent the states of the search in E-space. Using completed pdags instead of general pdags (or dags in the case of B-space) eliminates the problem of having multiple representations for the same equivalence class. We emphasize this result with a lemma.

**Lemma 2** *Let $\mathcal{P}^c{}_1$ and $\mathcal{P}^c{}_2$ denote two completed pdags that both admit a consistent extension. Then $\mathcal{P}^c{}_1 = \mathcal{P}^c{}_2$ if and only if $Class(\mathcal{P}^c{}_1) = Class(\mathcal{P}^c{}_2)$.*

**Proof:** Follows immediately by Theorem 1 and by the definitions of compelled and reversible edges.    □

To complete the specification of E-space, we define a set of simple operators that can be applied to completed pdags. The operators make extensive use of two algorithms. The first algorithm, which we refer to as PDAG-TO-DAG, takes as input an arbitrary pdag and returns a consistent extension if one exists. PDAG-TO-DAG returns an error message if the given pdag does not admit a consistent extension. Dor and Tarsi (1992) present an efficient implementation of PDAG-TO-DAG that we used for our experimental results. The second algorithm, which we refer to as DAG-TO-PDAG, takes as input an arbitrary dag $\mathcal{G}$ and returns the completed pdag representation for $Class(\mathcal{G})$. Chickering (1995b) , Meek (1995) , and Anderson et al. (1995) have independently derived implementations of DAG-TO-PDAG. For our experimental results, we used the algorithm derived by Chickering

(1995b), which is asymptotically optimal in the average case.

Given a completed pdag $\mathcal{P}^c$, we define the following four types of operators:

1. For any undirected edge $x - y$ in $\mathcal{P}^c$, we can delete the edge

2. For any directed edge $x \to y$ in $\mathcal{P}^c$, we can either delete the edge or reverse the edge.

3. For any pair of nodes $x$ and $y$ that are not adjacent in $\mathcal{P}^c$, we can either insert an undirected edge between $x$ and $y$, or insert a directed edge in either direction

4. For any triple of nodes $x$, $y$ and $z$ in $\mathcal{P}^c$, if $x$ and $z$ are not adjacent, and either $x$ and $y$ or $z$ and $y$ are adjacent, we can insert the v-structure $(x, y, z)$

All operators are subject to the constraint that the resulting pdag is acyclic[4] and admits a consistent extension. The operators are *complete* for the search space. That is, given any pair of completed pdags $\mathcal{P}^c{}_1$ and $\mathcal{P}^c{}_2$ that both admit a consistent extension, there exists a sequence of legal operators that moves from $\mathcal{P}^c{}_1$ to $\mathcal{P}^c{}_2$. A proof that the operators are complete can be found in [Anderson et al., 1995].

For a given completed pdag, let $\mathcal{P}$ denote the pdag—not necessarily completed—that results after directly applying one of the operators. The completed pdag that results from the operator is obtained as follows. First, the algorithm PDAG-TO-DAG is called with input $\mathcal{P}$ to extract a consistent extension $\mathcal{G}$. If $\mathcal{P}$ does not admit a consistent extension, then the given operator is not legal. To complete the application, the algorithm DAG-TO-PDAG is called with input $\mathcal{G}$ to build the resulting completed pdag representation. The process of applying an operator is depicted schematically in Figure 6, and the application of each operator type is illustrated in Figure 7.

Note that the consistent extension obtained during the application of an operator can be used to score the resulting state if the scoring function takes a dag as input, which is typically the case. Alternatively, PDAG-TO-DAG can be called with the completed pdag representation of a state whenever that state needs to be scored.

The proposed operators are all simple and local changes to the the edges in a pdag, but as we see in the example of Figure 6, an operator can have "cascading" effects on the pdag. Furthermore, a local change in a completed pdag may not correspond to a local change in the consistent extension used to score the equivalence class. This fact is unfortunate, because most scoring functions can exploit the locality of changes to dags to efficiently update the corresponding score. Fortunately, we have found that the cascading effects are uncommon in practice. For most of the operator applications, the resulting state can be scored by applying a small number of local changes to a consistent extension of the current state.

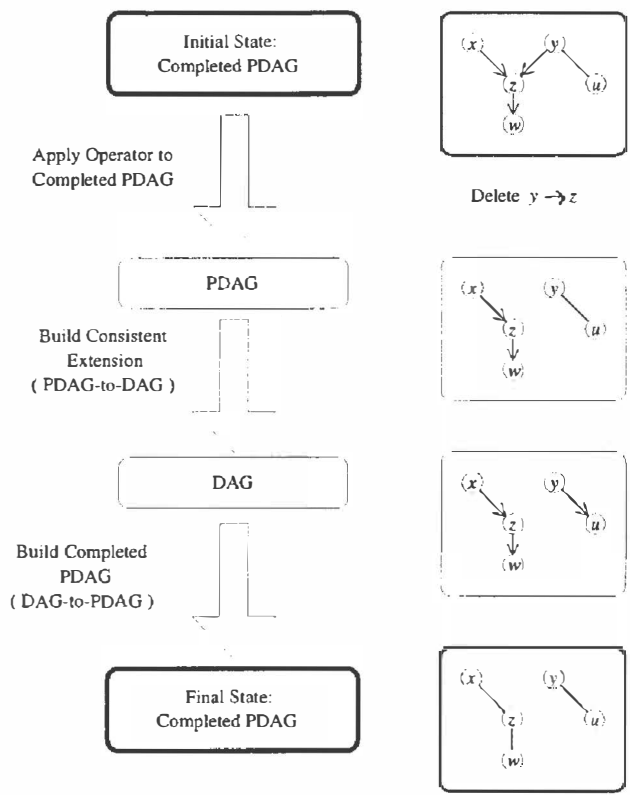

Figure 6: Diagram depicting how operators are applied. The pdags on the right give an example for every stage of the process.

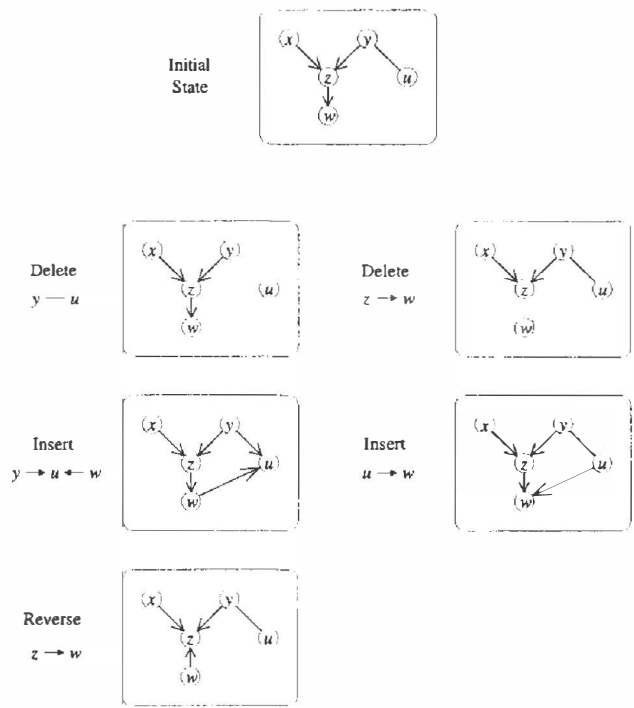

Figure 7: Example of each type of operator.

## 4    EXPERIMENTAL RESULTS

In this section, we compare the performance of greedy search when applied to E-space to the performance of greedy search when applied to B-space. Greedy search is a simple algorithm that, given the current state, always moves to the adjacent state that increases the score the most. If no adjacent state has a higher score, the algorithm terminates. In all of our experiments, we used greedy search initialized with the empty graph to identify a single equivalence class with a high score.

Given a particular scoring function, the goal of the search algorithm is to identify the structure (or structures) with the highest score. Consequently, the score of the best state found by the search algorithm is the best indicator of the algorithm's performance. Many researchers have compared scoring functions and search algorithms together. In this case, learning accuracy is often measured in terms of how close the learned network (or equivalence class) is to the *gold standard network*, or the model from which the train-ing data was generated. Typical choices for this comparison include structural difference and cross entropy. To be consistent with previous work, we have included a structural comparison in our experiments.

In our experiments, we compared the structure of the learned equivalence class to the structure of the equivalence class to which the gold standard belongs. To make this comparison, we first generated the completed pdag representation for both equivalence classes. Then, for each pair of nodes in the pdags, if the edge between those nodes were different in the two structures, we added one to the structural difference.

For all of our experiments, we used a score-equivalent special case of the *BDe* scoring function derived by Heckerman at al. (1995). Given data and prior knowledge, the BDe score is the relative posterior probability of the candidate structure using some reasonable assumptions. The version we used corresponds to an empty prior network and an equivalent sample size of either 8 (for the random graphs) or 16 (for the Alarm network). In all of the results reported, the BDe scores are expressed in log-space.

For each experiment, we present (1) the difference in score, (2) the difference in structural difference from the gold standard, and (3) the ratio of learning times. The difference in score is the score in E-space minus the score in B-space, and hence any positive number indicates that search in E-space is out-performing search in B-space. The difference in structural difference, however, is the structural difference in B-space minus the structural difference in E-space. We chose to present the results this way so that a positive difference in structural score also denotes a win for E-space. The ratio of times denotes how many times longer the search in E-space took than the search in B-space. This ratio is always greater than one.

Table 1: Effect of gold-standard size on search performance. Each score, structural difference, and time is the average over 9 total random databases consisting of 500 cases each.

| Number of Nodes | E-space Score | B-space Score | Score Diff. | E-space Struct | B-space Struct | Struct Diff | E-space Time (sec) | B-space Time (sec) | Time Ratio |
|---|---|---|---|---|---|---|---|---|---|
| 5 | -1326.64 | -1327.06 | 0.42 | 0.78 | 1.44 | 0.66 | 1 | 0 | — |
| 10 | -2745.55 | -2764.05 | 18.5 | 4.44 | 10.56 | 6.12 | 18.11 | 1.67 | 10.84431 |
| 15 | -3665.29 | -3677.17 | 11.88 | 17.67 | 21.89 | 4.22 | 70.44 | 6.22 | 11.32476 |
| 20 | -5372.94 | -5408.67 | 35.73 | 25.11 | 30.78 | 5.67 | 184.67 | 11.78 | 15.67657 |
| 25 | -6786.83 | -6860.24 | 73.41 | 32.67 | 47.11 | 14.44 | 487.33 | 22.56 | 21.60151 |

For many of the experiments we used randomly-generated gold standards. All random gold standards contained binary variables, and were generated as follows: for every pair of nodes in the graph, a directed edge was inserted with probability 0.3, subject to the constraint that no node could have more than 4 parents. Each conditional parameter set of the resulting network was drawn from a uniform Dirichlet distribution.

Our first experiment investigated how search performance in E-space compares to search performance in B-space as the size of the graphs increase. We used five sizes for the gold-standard in this experiment: 5, 10, 15, 20, and 25 nodes. For each graph size, we generated three random gold standards. For each gold standard, we generated three random databases, consisting of 500 cases each. We ran greedy search in both E-space and B-space, using each of the databases generated. The results are summarized in Table 1. For every gold-standard size, greedy search in E-space outperformed greedy search in B-space, both in terms of the average score and in terms of the average structural difference. Furthermore, as the complexity of the gold standard increased, the difference between the quality measures tended to increase as well.

Our next experiment compared the search spaces when the size of the database increases. We used random gold standards for this experiment, where each gold standard contained 10 binary nodes. We used six database sizes: 500, 1000, 1500, 2000, 2500, and 3000 cases. For each database size, three gold standards were generated; and for each gold standard, three databases containing the given number of cases were generated. For each database, greedy search was run in both E-space and B-space. The results are summarized in Table 2. Again, greedy search in E-space outperformed greedy search in B-space, in both quality measures, for every database size.

The final experiment was run using data generated from the Alarm network (Beinlich et al., 1989). The network, which contains 37 nodes and 46 edges, is an expert system for the problem of ICU ventilator management and has become a standard benchmark for learning algorithms. We generated 10 databases from the Alarm network, where each database contained 10000 cases. For each database, greedy search was run in both E-space and B-space. The results are summarized in Table 3. Greedy search in E-space significantly outperformed greedy search in B-space for this domain.

## 5 DISCUSSION

Overall, our empirical results show that the greedy algorithm applied to E-space consistently outperforms the greedy algorithm applied to B-space, although the time to complete the search in E-space was significantly longer. There are two reasons for the latter observation. Consider a search where all networks have $n$ nodes. In B-space, there are approximately $n(n-1)$ operators for greedy search to consider at each step. In E-space, however, there are approximately $n(n-1) + 2e(n-2)$ operators at each step, where $e$ denotes the number of edges in the current completed pdag (the extra operators correspond to v-structure insertions). In addition to the extra operators, there is also additional overhead to apply each operator. In particular, the algorithm PDAG-TO-DAG runs in time $O(n^2)$, and the algorithm DAG-TO-PDAG runs in time $O(e)$. Note that the extra overhead from these algorithms is a function of the network and does not depend on the size of the database. Consequently, as the data-sets grow large this overhead becomes less significant, as is demonstrated in Table 2.

One problem with the choice of greedy search for our comparisons is that the algorithm does not benefit from additional time once a local maximum is reached. Consequently, we do not have a comparison of E-space and B-space for algorithms that are given the same time constraint. Such a comparison would be useful if more sophisticated search algorithms are going to be applied to the learning problem.

An interesting extension to this work is to combine E-space and B-space. One approach would be to run the greedy search in B-space until a local maximum is reached. Next, generate the completed pdag representation and see if the score can be increased in E-space. If it can, make one step in E-space and then switch back to B-space and go to the next local maximum. This approach will be fast because each local maximum is reached in B-space. Furthermore, by using E-space to get out of these local maxima, the resulting search performance may improve.

Table 2: Effect of database size on search performance. Each score, structural difference, and time is the average over 9 total databases generated from gold standards containing 10 nodes.

| DB Size | E-space Score | B-space Score | Score Diff. | E-space Struct | B-space Struct | Struct Diff | E-space Time (sec) | B-space Time (sec) | Time Ratio |
|---------|---------------|---------------|-------------|----------------|----------------|-------------|--------------------|--------------------|------------|
| 500     | -2745.55      | -2764.05      | 18.5        | 4.44           | 10.56          | 6.12        | 18.11              | 1.67               | 10.84431   |
| 1000    | -5399.82      | -5449.35      | 49.53       | 2.67           | 9.22           | 6.55        | 21.67              | 3.33               | 6.507508   |
| 1500    | -8092.83      | -8148.43      | 55.6        | 3.56           | 8.22           | 4.66        | 25.67              | 4.33               | 5.928406   |
| 2000    | -10724.9      | -10825.1      | 100.2       | 2              | 7              | 5           | 34.56              | 6.11               | 5.656301   |
| 2500    | -13386.5      | -13416.6      | 30.1        | 2.78           | 10.78          | 8           | 36.89              | 8.22               | 4.487835   |
| 3000    | -16050.7      | -16146        | 95.3        | 2.78           | 12.22          | 9.44        | 43                 | 9.33               | 4.608789   |

Table 3: Greedy search performance for the Alarm network. Scores, structural differences, and times are averages over 10 databases, where each database contains 10000 cases.

| E-space Score | B-space Score | Score Diff. | E-space Struct | B-space Struct | Struct Diff | E-space Time (sec) | B-space Time (sec) | Time Ratio |
|---------------|---------------|-------------|----------------|----------------|-------------|--------------------|--------------------|------------|
| -101004       | -101255       | 250.8       | 36.3           | 51.5           | 15.2        | 10503.1            | 526.2              | 19.96      |

## Acknowledgments

I would like to thank David Heckerman, Rich Korf, David Madigan, Chris Meek, and anonymous reviewers for useful suggestions. This work was supported by NSF Grant No. IRI-9119825, and a grant from Rockwell International.